\documentclass{article}

\usepackage{arabtex}

\usepackage[final,nonatbib]{neurips_2024}

\usepackage[utf8]{inputenc} 
\usepackage[T1]{fontenc}    
\usepackage{hyperref}       
\usepackage{url}            
\usepackage{booktabs}       
\usepackage{amsfonts}       
\usepackage{nicefrac}       
\usepackage{microtype}      
\usepackage{xcolor}         

\usepackage{soul}
\usepackage{utf8}

\title{CamelEval: Advancing Culturally Aligned Arabic Language Models and Benchmarks}

\author{%
  Zhaozhi Qian \\
  Elm Company\\
  \texttt{zqian@elm.sa} \\
  \And
  Faroq Altam \\
  Elm Company \\
  \texttt{faltam@elm.sa} \\
  \AND
  Muhammad Alqurishi \\
  Elm Company \\
  \texttt{mualqurishi@elm.sa} \\
  \And
  Riad Souissi \\
  Elm Company \\
  \texttt{rsouissi@elm.sa} \\
}

\begin{document}
\setcode{utf8}

\maketitle

\begin{abstract}
Large Language Models (LLMs) are the cornerstones of modern artificial intelligence systems. This paper introduces Juhaina, a Arabic-English bilingual LLM specifically designed to align with the values and preferences of Arabic speakers. Juhaina inherently supports advanced functionalities such as instruction following, open-ended question answering, information provisioning, and text processing. Our model contains 9.24 billion parameters and is trained on a context window of up to 8,192 tokens. This paper details the creation process of Juhaina and provides an extensive empirical evaluation. Furthermore, we identify the limitations of widely-adopted Open Arabic LLM Leaderboard (OALL) and propose a new evaluation benchmark, CamelEval. Our findings demonstrate that Juhaina surpasses existing LLMs of comparable sizes, such as the Llama and Gemma families, in generating helpful responses in Arabic, providing factually accurate information about the region, and understanding nuanced cultural aspects. We aspire for Juhaina to democratize cutting-edge AI technologies, serving over 400 million Arabic speakers by offering LLMs that not only communicate in their language but also comprehend their culture. We publicly release all models on Huggingface\footnote{The Juhaina LLMs can be downloaded from \url{https://huggingface.co/elmrc}}.
\end{abstract}

\section{Introduction}

Large Language Models (LLMs) have emerged as transformative tools in artificial intelligence. LLMs are believed to significantly enhance productivity by automating complex tasks such as content creation and customer service \cite{brown2020language}. Moreover, they enable more intuitive and accessible interactions with technology, from personalized virtual assistants to advanced language translation services. 

Currently, approximately 400 million people worldwide speak Arabic \cite{Ara}. However, Arabic content constitutes only a small fraction of the web corpus used for pre-training large language models (LLMs). For example, Common Crawl contains merely 0.6\% of documents in Arabic \cite{CC}. This scarcity is compounded by the unique linguistic patterns and the rich cultural and historical connotations inherent in the Arabic language. Consequently, many existing Arabic-centric \cite{sengupta2023jais,huang2023acegpt} or multilingual LLMs \cite{aryabumi2024aya,penedo2023refinedweb,muennighoff2022crosslingual} struggle to generate responses that are both useful and culturally aligned in Arabic \cite{naous2023having,koto2024arabicmmlu}.


To better serve the Arabic-speaking community, we set the goals as follows:

\textbf{1. Proficiency in the Arabic Language.} The LLM should understand user inputs in Arabic, generating coherent and grammatically correct outputs. It should avoid translationese, producing natural, human-like responses. Additionally, the LLM should effectively articulate technical topics using domain-specific terminology, such as in science.

\textbf{2. Awareness of Local Facts.} The LLM should be knowledgeable about facts commonly known in Arabic-speaking regions, including history and geology. It should provide accurate information without hallucination and incorporate this knowledge into general conversations.

\textbf{3. Alignment with Arabic Culture.} The LLM should generate responses that are appropriate and respectful to the Arabic audience, adhering to the cultural and social norms of the region.

In this paper, we introduce Juhaina, a novel Arabic-English bilingual LLM, alongside CamelEval, a new benchmarking tool, marking strides toward achieving these objectives. Juhaina is a decoder-only dense transformer model, trained on context windows extending up to 8192 tokens, and has 9.24 billion parameters. This design allows for inference and further adaptation without the need for high-end GPUs, enhancing its accessibility to the broader community. Juhaina is post-trained on Gemma 2, a series of open-weight LLMs made available by Google \cite{team2024gemma}.


CamelEval is a new benchmark for Arabic LLMs, specifically designed to assess their conversational abilities and instruction-following proficiency within Arabic contexts. This benchmark embraces the llm-as-a-judge framework, an approach pioneered by AlpacaEval\cite{alpaca_eval}. To evaluate the LLMs' capability to produce coherent responses that accurately reflect cultural nuances and regional facts, we have carefully trans-curated two sets of challenging prompts. We aspire that CamelEval would play a crucial role in gauging progress towards achieving the three key goals.

In this report, we outline the development and assessment of Juhaina, aiming to contribute our insights to the broader community. We delve into our process of data curation, which involves systematically searching for, collecting, annotating, and refining datasets. This rigorous approach yields a high-quality dataset that integrates translated open-source data, synthetic data, and human annotations. Furthermore, we discuss our post-training strategies, which encompass supervised fine-tuning (SFT) and adjustments based on human feedback.


We conducted a thorough evaluation of Juhaina, employing a wide array of benchmarks to gain a comprehensive understanding of its performance. Our evaluation incorporated benchmarks from the Open Arabic LLM Leaderboard (OALL) \cite{OALL} and CamelEval. The results of our assessment show that Juhaina surpasses other LLMs of similar size in delivering useful responses within Arabic contexts. This success represents a crucial advancement towards our goals of developing a model that not only demonstrates a deep understanding of the Arabic language but also effectively captures and aligns with the cultural and contextual nuances specific to Arabic-speaking regions.

We publicly release all Juhaina LLMs under MIT license. We aim for this release to serve the Arabic-speaking communities and help narrow the disparity in accessing advanced AI technology.

\section{Creating Juhaina LLMs}
\label{creation}

Juhaina is created by performing rigorous post-training on the Gemma 2 family of models, which were pre-trained on up to 13 trillion tokens of primarily-English data \cite{team2024gemma}. 
Our post-training process comprises two stages: the supervised fine-tuning (SFT) stage and the alignment with human preference stage.

In the SFT stage, the Juhaina LLM is trained to reproduce responses crafted by human annotators. This involves exposing the model to a diverse set of high-quality, human-generated text examples, ensuring it learns to generate coherent and contextually appropriate responses. The goal here is to refine the model's understanding of language and improve its ability to generate text that closely mirrors human communication.

In the alignment with human preference stage, the LLM is further refined to adapt its style and tone to align with human preferences. This stage involves using techniques such as reinforcement learning from human feedback (RLHF), where the model's outputs are evaluated and ranked by human reviewers. The model is then adjusted based on this feedback to better match the desired style, tone, and appropriateness of responses. This ensures that the LLM not only produces accurate and contextually relevant information but also communicates in a manner that is consistent with human expectations and cultural norms.

\subsection{Data Collection}
\label{sec:data_collection}

Collecting Arabic datasets was one of the main bottlenecks for developing Juhaina. Most of the open Arabic datasets are translated from other languages and are subject to translation biases or fail to reflect cultural context appropriately \cite{vanmassenhove2021machine,stanovsky2019evaluating,wang2022measuring,naous2023having}. Moreover, regional datasets are extremely scarce and with limited quality \cite{singh2024aya,alyafeai2024cidar}.
    

\subsubsection{Data Sources}

To systematically collect data, we first identified a comprehensive list of subject categories, such as ``Science'' and ``Humanities.'' For each subject, we further delineated sub-categories, such as ``Physics'' and ``History.'' For each identified category, we conducted a thorough search to locate and collect relevant datasets. Our search methodology encompassed the following data sources:

\textbf{1. Internal Datasets:} We identified existing datasets within our database. These datasets were evaluated for relevance and quality to ensure they met our requirements.

\textbf{2. Open Web Search:} We performed extensive searches on the open web to find websites containing useful data that could be scraped. This involved using advanced search techniques and tools to identify and extract data from various online sources. The data obtained from these sources was then cleaned, validated, and integrated into our dataset.

\textbf{3. Translatable Datasets:} We located datasets available in different languages. These datasets were flagged for translation to ensure their effective utilization within our project. The translation process involved both automated tools and manual verification to maintain accuracy.

During the search, we considered a variety of criteria, such as relevance, accuracy, and timeliness. The full list of criteria is available in the Appendix. We used a tagging system to annotate various aspects of the data and flagged data with high uncertainty for human review. We prioritized Modern Standard Arabic (MSA) data, ensured sources were reputable, and preferred cleaned data. We collected the resulting dataset into a corpus with the associated category tags for further processing. 

\subsubsection{Data Cleaning}

To ensure the quality and relevance of our dataset, we implemented a multi-step data cleaning process. Initially, we applied basic filters to eliminate empty and invalid rows, which helped in removing obvious errors and gaps in the data. Following this, we employed advanced filtering techniques, including semi-regular expressions (semi-regex) and fuzzy matching. These methods allowed us to perform more nuanced cleaning, such as removing unwanted characters like emojis and filtering out non-Arabic data records, which could otherwise introduce noise and inaccuracies. We have also take care to avoid data contamination by removing data that duplicate or paraphrase the benchmarks. 

\subsubsection{Prompt Generation}

We defined a list of LLM capabilities that we would like Juhaina to achieve, such as factuality and reasoning (the full list of capabilities is provided in the Appendix). For each document in the data corpus, the annotators generated prompts that invoke different LLM capabilities. Each prompt may contain a context, which is an extract or summary of the original document, followed by the task instruction or question. In some cases, the annotator also provided a ``system prompt'' that specifies the role of the LLM and the expected style of the response.

After curating the initial set of prompts, we used an automated tool to tag the quality and difficulty of each prompt. Prompts with low quality are often ambiguous and lack clear instructions. We removed these low-quality prompts from the final set.

\subsubsection{Answer Generation}

We split the set of prompts into two parts: one for supervised fine-tuning (SFT) and the other for alignment through human feedback. For the SFT data, the annotators provided the ground truth responses to the prompts. For the alignment data, the annotators provided preference feedback on pairs of answers generated by the LLM.

\subsubsection{Postprocessing}

The dataset then undergoes three postprocessing pipelines: \textit{filtering}, \textit{transformation}, and \textit{balancing}.

The filtering pipeline cleans the data by removing duplicates, unnecessary white spaces, foreign language content, and other irrelevant elements. It also verifies the overlap between each sample and the evaluation data to ensure there is no contamination with the test set. The transformation pipeline converts the cleaned data into the desired alignment format. During each stage, native speakers conduct manual investigations to address any non-trivial cases.

Once the data is processed, a further level of re-sampling is performed to improve data distribution. We sub-sample an equal number of rows for each subject category, resulting in a final dataset comprising approximately 600,000 rows for SFT and 40,000 rows for alignment.


\subsubsection{Learnings on Data Collection}

We have learned several key lessons from our data curation process. 

\textbf{Focus on Data Quality}. There is a well-known trade-off between quality and quantity in data collection. We found that quality matters much more than quantity in post-training. In fact, our experience shows that including low-quality data significantly hurts benchmark performance. However, in practice, we tend to overemphasize data quantity simply because it is easier to measure (i.e., just \textit{count} the tokens or rows), whereas understanding data quality is quite non-trivial. Based on the problems we have encountered and resolved during the curation process, we created a checklist of common data quality issues with multi-lingual datasets:


\begin{table}[t!]
\centering
\begin{tabular}{@{}l@{}}
\toprule
Data Quality Checklist                        \\ \midrule
Duplicate or near-duplicate data                     \\
Missing data and other artifacts due to web scraping \\
Artifacts introduced by translation                  \\
Data in out-of-scope languages                       \\
Ill-formatted code blocks or structured text         \\
Irrelevant system prompts                            \\
Unbalanced mix of tasks, categories, or difficulty   \\ \bottomrule
\end{tabular}
\vspace{2pt}
\caption{Common data quality issues with multi-lingual datasets.}
\end{table}

\textbf{Tag the Data}. 
We found that extensively tagging the data greatly facilitates the curation process. It not only exposes data quality issues but also informs data selection and sampling. We adopted a variety of automated tagging systems and human annotations to understand various aspects of the data, including subject category and sub-category, task type, task difficulty, and more.

\textbf{Effective Communication with Annotators}. We extensively used human annotation and feedback in the curation process. We found that effective communication with the annotators is vital for success. Care needs to be taken to ensure that the annotators understand and adhere to the desired annotation practices. Providing practical examples and a walk-through of the process helps align expectations and behaviors. Failure or delay in communication could lead to systematic errors or biases in the data, rendering the effort worthless.

\subsection{Post-training Procedure}

\subsubsection{Supervised Finetuning (SFT)}
\label{sft}


We used the Llama Pro algorithm \cite{wu2024llama} for supervised fine-tuning (SFT). Llama Pro performs block expansion on the base model to introduce additional free parameters. During the training stage, only these newly added parameters are trained. Our experiments indicate that, compared to standard SFT, Llama Pro enables the LLM to acquire additional capabilities in the Arabic language without compromising its existing capabilities in English. The training settings are detailed in Table \ref{tbl:hyperparams}.

\begin{table}[!th]
    \centering
    \begin{tabular}{lll}
        \toprule
        {Parameter} & {SFT} & {Align.} \\ \midrule
        Optimizer & AdamW & AdamW \\
        Method & Llama pro \cite{wu2024llama} & Freeze \\
        Initial lr & 5e-5 &  1e-5 \\
        Scheduler & Cosine & Cosine \\
        Epochs  & 3 & 1\\
        GPU (per node)  & 8 & 8 \\
        Nodes      & 8 & 8 \\
        Batch size  & 64 &32  \\
        \bottomrule
    \end{tabular}
    \caption{\label{tbl:hyperparams} The parameters being adopted for training.}
\end{table}

\subsubsection{Alignment with Human Feedback}
\label{dpo}
At the conclusion of the Supervised Fine-Tuning (SFT) stage, the Juhaina model has demonstrated proficiency in following instructions and adapting to domain-specific knowledge. However, additional post-processing is necessary to ensure safety and adherence to desired behaviors, such as compliance with specific cultural norms.

This requirement is particularly important given the flexibility of the Arabic language and the diversity of its dialects and cultural contexts. To address these challenges, we have implemented a bespoke data curation approach aimed at achieving high data quality. In particular, the alignment data is meticulously collected and validated by native Arabic speakers to ensure accuracy and cultural relevance. 

We used the Monolithic Odds Ratio Preference Optimization (ORPO) algorithm \cite{hong2024orpo} to align Juhaina with human preference data. ORPO is inspired by the Direct Preference Optimization (DPO) algorithm \cite{rafailov2024direct}. Based on empirical evidence from our pilot studies, we selected ORPO for two primary reasons. Firstly, unlike DPO, ORPO does not require a reference model, which significantly reduces GPU memory usage during training. Secondly, ORPO naturally incorporates a supervised fine-tuning (SFT) loss, which is known to stabilize the training process \cite{dubey2024llama}.

Similar to the SFT stage, we adopted freeze tuning, adapting only a subset of the model parameters. Our pilot studies indicated that freeze tuning achieves better performance compared to full parameter tuning or other parameter-efficient methods, such as LoRa \cite{hu2021lora}. The training parameters are detailed in Table \ref{tbl:hyperparams}.
The computing setup is the same as the SFT stage.

\subsubsection{Infrastructure and Computing}

The SFT was conducted on our computing cluster, which consists of 8 nodes, each equipped with 8 GPUs featuring 80GB of VRAM. We utilized PyTorch's Fully Sharded Data Parallel (FSDP) to efficiently manage the parallel training process. The training was completed in 3 days.

\section{Evaluation of Arabic LLMs} \label{eval}

\subsection{Open Arabic LLM Leaderboard (OALL)} \label{eval-oall}

The Open Arabic LLM Leaderboard (OALL) is one of the few close-ended Arabic benchmarks \cite{OALL}. It encompasses three primary benchmarks: AlGhafa \cite{almazrouei-etal-2023-alghafa}, ACVA \cite{ACVA}, and Arabic MMLU \cite{koto2024arabicmmlu}, which are all curated specifically for Arabic LLMs. OALL also include translated versions of standard LLM benchmarks such as EXAMS, ARC, BOOLQ, COPA, HELLASWAG, OPENBOOK-QA, PIQA, RACE, SCIQ, and TOXIGEN. They assess the model's capability in selecting the correct answer from \textit{yes-no} or multiple-choice options in the zero-shot setting. The normalized log-likelihood of the target is used to measure the model's ability to generate the correct choice. The OALL has a centralized and managed process to evaluate all submitted LLMs. This ensures reproducibility and standardization of the results. 

The latest leaderboard is available online \footnote{\url{https://huggingface.co/spaces/OALL/Open-Arabic-LLM-Leaderboard}}. In Table \ref{tab:oall}, we reproduce the results of the top-five LLMs among those with less than 13B parameters. We note that Juhaina is currently at the second place, achieving strong performance in the overall average score and the three main Arabic benchmarks. Remarkably, Silma-9B-v1 stands out for its incredible performance, leading the overall leaderboard and surpassing models even with more than 70B parameters. 

OALL is instrumental in evaluating text completion capabilities, logical correctness, and factual knowledge across different domains. However, it has several key limitations: 

\textbf{1. Narrow coverage of LLM Capabilities.}
The benchmark's reliance on multiple-choice questions means it fails to evaluate the broader spectrum of LLM capabilities, such as engaging in conversations or following instructions. It does not measure the helpfullness or the utility of the LLM's replies, which are essential aspects of its performance. In fact, the ability to participate in general conversations is a defining feature of LLMs. 
Consequently, while this benchmark is effective for assessing the foundational knowledge and reasoning skills of pre-trained LLMs, it does not adequately measure the performance of LLMs that have been instruction-finetuned for generating meaningful interactions with users.

\textbf{2. Oversimplified evaluation metric.} The evaluation metric used by OALL, the normalized log-likelihood (NLL), is overly simplistic. NLL calculates the log-probability of producing the "gold response," adjusted for the length of this ideal response. However, the assumption that there's a singular "gold response" is flawed, even in contexts like multiple-choice questions. This inconsistency is apparent in OALL itself, where some correct answers are labeled as A, B, C, or D, and others are identified by the text of the correct option\footnote{Adding few-shot examples may help alleviate the arbitrariness of "gold response'', but OALL employees zero-shot evaluation in all cases \cite{OALL}.}. The variability in defining what constitutes a "gold response" renders NLL an unreliable and imprecise metric for LLMs, which can generate texts in diverse formats and styles.

\textbf{3. Public test set.} Publishing the benchmark test set enhances transparency, but it simultaneously opens the door for gaming the leaderboard. Specifically, individuals could fine-tune the LLM using a compromised dataset, leading to inflated scores. We have taken considerable efforts in our data curation process to minimize the chance of test set contamination (Section \ref{sec:data_collection}). However, considering the widespread recognition and credibility of OALL, it's plausible that some might be motivated to undertake deceptive practices.


\begin{table}[t!]
\centering
\begin{tabular}{@{}lcccccc@{}}
\toprule
Model         & Size (B) & Average & ACVA  & AlGhafa & AR-MMLU  & TOXIGEN \\ \midrule
Silma-9B-v1      & 9.24         & 70.62   & 78.89 & 71.85   & 52.55 & 67.59   \\
Gemma-2-SimPO & 9.24         & 58.55   & 43.45 & 54.2    & 53.65 & 82.46   \\
SauerkrautLM  & 12.25        & 57.35   & 39.54 & 56.09   & 43.99 & 79.04   \\
Gemma-2-SPPO  & 9.24         & 57.16   & 46.97 & 52.55   & 52.91 & 81.28   \\ \midrule
\textbf{Juhaina}    & 9.24         & 60.18   & 61.82 & 60.47   & 51.06 & 80.21   \\ \bottomrule
\end{tabular}
\vspace{2pt}
\caption{Top performing LLMs on the OALL with model sizes between 7B to 13B at the time of writing. The other models being compared are ``Silma-9B-v1''(\href{https://huggingface.co/silma-ai/SILMA-9B-Instruct-v1.0}{silma-ai/Silma-9B-Instruct-v1.0}
),  ``Gemma-2-SimPO''(\href{https://huggingface.co/princeton-nlp/gemma-2-9b-it-SimPO}{princeton-nlp/gemma-2-9b-it-SimPO}
), ``SauerkrautLM''(\href{https://huggingface.co/VAGOsolutions/SauerkrautLM-Nemo-12b-Instruct}{VAGOsolutions/SauerkrautLM-Nemo-12b-Instruct}
), and ``Gemma-2-SPPO'' (\href{https://huggingface.co/UCLA-AGI/Gemma-2-9B-It-SPPO-Iter3}{UCLA-AGI/Gemma-2-9B-It-SPPO-Iter3}
).}
\label{tab:oall}
\end{table}

\subsection{CamelEval}
\label{eval-alpaca}

To address the limitations in OALL, we developed a new evaluation benchmark, CamelEval. CamelEval is based on AlpacaEval \cite{alpaca_eval}, an LLM-as-a-judge automatic evaluation framework for instruction-following LLMs. Essentially, two competing LLMs provide responses to a set of test prompts, and they are subsequently evaluated by a ``judge'' LLM to decide the win rate. 

CamelEval enables a better coverage of LLM capabilities, especially the ability to generate helpful conversations and following user instruction. Using a LLM as a judge, CamelEval avoids much of the limitations of simple metrics, such as NLL. Furthermore, we included new prompts that have not appear in OALL before, reducing the chance of contamination. Therefore, we believe that CamelEval offers additional insights into LLM capabilities, which are more aligned with the practical use cases of instruction-following AI assistants.

CamelEval comprises two distinct test sets. The first, known as the Translated Set, consists of translations of the 805 prompts found in the original AlpacaEval. The English version of these prompts has gained widespread use for evaluating LLMs and has demonstrated a high level of concordance with assessments made by human evaluators. To assemble the Translated Set, we enlisted native-speaking annotators to refine the initial translations provided by Freedom Intelligence (\href{https://huggingface.co/datasets/FreedomIntelligence/Arabic-AlpacaEval/tree/main}{link}). These annotators have successfully identified and corrected various anomalies that likely originated from the process of machine translation.

To better reflect Arabic cultural linguistic nuances, we developed the second test set, named the Curated Set. This premium test set contains 805 prompts, synthetically created using GPT-4, based on a corpus of human-curated, textbook-quality content spanning various fields such as culture, religion, history, sociology, and other scientific disciplines. To ensure the prompts encapsulate complex instructions and pose challenging questions, we directed GPT-4 to formulate questions that go beyond the information provided in the text, necessitating supplementary knowledge for accurate responses. We then assessed the difficulty level of these questions, selecting only the most challenging for inclusion. To guarantee the accuracy and relevance of these prompts, annotators manually reviewed each one. The specific subject categories covered in the Curated Set are detailed in Appendix \ref{app:subject}.

In Table \ref{tab:camel-eval}, we present a comparison of Juhaina's performance against a selection of LLMs that are either focused on Arabic or are multilingual, and that have a parameter count equal to or larger than that of Juhaina. For this comparison, Gemma2-9b-IT serves as the baseline against which the win rate is calculated.

We observe that in the Translated Set, Juhaina's performance is on par with that of the much larger 70B Qwen2-Instruct LLM, showcasing its efficiency in handling translated prompts. In the more demanding Curated Set, Juhaina closely follows the performance of GPT-4o, indicating its strong capability in addressing complex, culturally nuanced questions. Additionally, we note a significant and consistent underperformance by Silma-9B-v1 compared to the model it was fine-tuned from, Gemma2-9b-IT, with a win rate of less than 50\% across both test sets. This highlights a potential issue in the fine-tuning process or the adaptability of Silma-9B-v1 to the specific challenges presented by the CamelEval benchmark.

We note that the typical constraints associated with using LLMs as evaluators also apply to CamelEval. For example, there's a possibility that the judge LLM might show a preference for answers it generates itself. Additionally, other factors, like the length of the response, could introduce biases into the evaluation process. We aim to tackle these and other unresolved challenges in future updates of CamelEval.


\begin{table}[t!]
\centering
\begin{tabular}{@{}lccccc@{}}
\toprule
                  &              & \multicolumn{2}{c}{Curated} & \multicolumn{2}{c}{Translated} \\
   Model                  & Size (B) & Win Rate \%      & Std       & Win Rate \%       & Std        \\ \midrule
GPT-4o-2024-05-13           & -            & 99.73            & 0.11      & 97.11             & 0.50       \\
Qwen2-72B-Instruct          & 72.7         & 95.80            & 0.57      & 91.95             & 0.83       \\
MBZUAI-ORYX-new             & 43.7         & 93.55            & 0.72      & 85.70             & 1.09       \\
Gemma2-9B-IT                & 9.24         & 50.00            & 0.00      & 50.00             & 0.00       \\
Jais-Inception-70b & 72.7         & 43.67            & 1.68      & 43.08             & 1.66       \\
Silma-9B-v1                 & 9.24         & 30.56            & 1.49      & 45.26             & 1.61       \\ \midrule
Juhaina-9B                  & 9.24         & 98.47            & 0.36      & 91.37             & 0.88       \\ \bottomrule
\end{tabular}
\vspace{2pt}
\caption{Performance of Juhaina and other notable Arabic-centric or multilingual LLMs of equal or larger sizes on CamelEval. We report the winrate against the Gemma2-9B-IT model on the curated and translated prompt sets. Juhaina is able to generate more helpful responses for challenging Arabic queries compared to LLMs of 4-7 times the size.}
\label{tab:camel-eval}
\end{table}

\section{Discussion}

In this study, we have detailed the development of Juhaina, a leading open-weight Arabic LLM, and introduced CamelEval, a new benchmark for Arabic LLMs that serves as a complement to the existing OALL benchmarks. We aspire that the release of Juhaina's LLM weights, along with the CamelEval benchmark and the insights shared in this work, will assist the community in advancing the creation of improved and more culturally attuned LLMs for the Arabic-speaking world.



\bibliography{main}
\bibliographystyle{acm}

\newpage

\appendix

\section{Appendix}

\subsection{Criteria for Data Search}

\begin{enumerate}
    \item Relevance to Topic Criteria: The data must be directly related to our subject categories. 
    \item Timeliness Criteria: The data should be up to date. 
    \item Completeness Criteria: The dataset should be comprehensive enough to support robust analysis. 
    \item Granularity Criteria: The data should have the appropriate level of detail. 
    \item Availability and Accessibility Criteria: The data should be accessible and has an open license. 
    \item Bias and Objectivity Criteria: The data should be free from bias or, if biased, the bias should be understood and accounted for.
    \item Cost: The estimated cost for accessing and curating the data. 
\end{enumerate}

\subsection{List of LLM Capabilities}
\begin{enumerate} 
\item \textbf{Information Provision:} Delivering specific information or facts on various topics.

\item \textbf{Reasoning:} Engaging in logical thinking, problem-solving, or managing complex tasks.

\item \textbf{Planning:} Assisting in the creation of plans or strategies for activities and projects.

\item \textbf{Editing:} Editing, rephrasing, proofreading, or performing other tasks related to the composition of written content.

\item \textbf{Coding \& Debugging:} Assisting with writing, reviewing, or fixing code in programming.

\item \textbf{Math:} Providing help with mathematical concepts, problems, and calculations.

\item \textbf{Role Playing:} Interacting with users by adopting a character or persona.

\item \textbf{Data Analysis:} Interpreting data, analyzing statistics, or performing analytical tasks.

\item \textbf{Creative Writing:} Crafting stories, poems, or other creative texts.

\item \textbf{Advice Seeking:} Offering recommendations or guidance on various personal or professional issues.

\item \textbf{Brainstorming:} Generating ideas, engaging in creative thinking, or exploring possibilities.

\item \textbf{Other:} Addressing queries that do not fit into the above categories. 
\end{enumerate}

\subsection{Examples of Translation Issues in OALL}
\label{app:translation}

Example One:
\begin{RLtext}
انطفأت الشعلة الموجودة على الشمعة.، لأن :

0) لقد فجرت على الفتيل.

1) لقد وضعت عود ثقاب على الفتيل.

الإجابة:لقد فجرت على الفتيل
\end{RLtext}

Example Two:
\begin{RLtext}
شرب الرجل بكثرة في الحفلة.، لذلك :

0) كان يعاني من الصداع في اليوم التالي.

1) أصيب بسيلان في الأنف في اليوم التالي.

الإجابة:كان يعاني من الصداع في اليوم التالي.
\end{RLtext}

Example Three:
\begin{RLtext}
أخطأ الطالب في كتابة الكلمة.، لذلك :

0) قام المعلم بتصحيحها.

1) طردتها المعلمة .

الإجابة:
\end{RLtext}

Example Four:
\begin{RLtext}
امرأة ببدلة حمراء تتحدث خلف مكتب. رجل
الاقتراحات:

0) يأخذ كرات من حامل على الجدار.

1) يقوم بتنظيف كرسي أسود بالمكنسة الكهربائية.

2) يرش الماء في حوض أبيض صغير.

3) يقف ويحمل قدمه خلف ظهره.

الإجابة:يقف ويحمل قدمه خلف ظهره
\end{RLtext}

\subsection{Subject Categories in CamelEval Curated Set}
\label{app:subject}

\begin{table}[ht]
\centering
\begin{tabular}{@{}lc@{}}
\toprule
Category      & Occurrence \\ \midrule
Physics       & 66        \\
Biographic    & 65        \\
Economy       & 62        \\
Philosophy    & 61        \\
Geology       & 61        \\
Psychology    & 60        \\
Nutrition     & 59        \\
Chemistry     & 58        \\
Education     & 56        \\
Arts          & 54        \\
Medicine       & 53        \\
Arabic        & 50        \\
Math          & 39        \\
Engineering   & 24        \\
Tech          & 16        \\
Anthropology  & 14        \\
Infographic   & 3         \\
Environment & 2         \\
History       & 2         \\ \bottomrule
\end{tabular}
\vspace{2in}
\caption{Subject categories in CamelEval Curated Set. The prompts in the Curated Set covers a wide range of topics that are science and culture related.}
\end{table}

\end{document}